%% file: main.tex
\documentclass[letterpaper]{article} 
\usepackage{aaai25}  
\usepackage{times}  
\usepackage{helvet}  
\usepackage{courier}  
\usepackage[hyphens]{url}  
\usepackage{graphicx} 
\def\UrlFont{\rm}  
\usepackage{natbib}  
\usepackage{caption} 
\usepackage[linesnumbered,ruled]{algorithm2e}

\usepackage{caption}
\usepackage{kotex} 
\usepackage{soul} 
\usepackage{multirow}
\usepackage{adjustbox}
\usepackage{subcaption}
\usepackage{dsfont}
\usepackage{bm}
\usepackage{amsmath}
\usepackage{cleveref}
\usepackage{amssymb}
\usepackage{booktabs}
\usepackage{comment}
\usepackage{pifont}
\newcommand{\cmark}{\ding{51}}%
\newcommand{\xmark}{\ding{55}}%

\usepackage{kotex}
\usepackage[usenames,dvipsnames]{color}

\usepackage{xcolor}

\title{DomCLP: Domain-wise Contrastive Learning with Prototype Mixup\\for Unsupervised Domain Generalization}
\author{
    Jin-Seop Lee, Noo-ri Kim, Jee-Hyong Lee\thanks{Corresponding author}
}
\affiliations{
    Sungkyunkwan University, Suwon, South Korea\\
    \{wlstjq0602, pd99j, john\}@skku.edu
}

\usepackage{bibentry}

\begin{document}
\maketitle
\input{CameraReady/LaTeX/sections/0_abstract}
\input{CameraReady/LaTeX/sections/1_introduction}
\input{CameraReady/LaTeX/sections/2_related}
\input{CameraReady/LaTeX/sections/3_proposed}
\input{CameraReady/LaTeX/sections/4_experiments}

\input{CameraReady/LaTeX/sections/5_conclusion}
\bibliography{aaai25}
\input{CameraReady/LaTeX/sections/6_appendix}
\end{document}

%% file: CameraReady/LaTeX/sections/0_abstract.tex
\begin{abstract}
Self-supervised learning (SSL) methods based on the instance discrimination tasks with InfoNCE have achieved remarkable success. Despite their success, SSL models often struggle to generate effective representations for unseen-domain data. To address this issue, research on unsupervised domain generalization (UDG), which aims to develop SSL models that can generate domain-irrelevant features, has been conducted. Most UDG approaches utilize contrastive learning with InfoNCE to generate representations, and perform feature alignment based on strong assumptions to generalize domain-irrelevant common features from multi-source domains. However, existing methods that rely on instance discrimination tasks are not effective at extracting domain-irrelevant common features. This leads to the suppression of domain-irrelevant common features and the amplification of domain-relevant features, thereby hindering domain generalization. Furthermore, strong assumptions underlying feature alignment can lead to biased feature learning, reducing the diversity of common features. In this paper, we propose a novel approach, \textit{DomCLP}, Domain-wise Contrastive Learning with Prototype Mixup. We explore how InfoNCE suppresses domain-irrelevant common features and amplifies domain-relevant features. Based on this analysis, we propose Domain-wise Contrastive Learning (\textit{DCon}) to enhance domain-irrelevant common features. We also propose Prototype Mixup Learning (\textit{PMix}) to generalize domain-irrelevant common features across multiple domains without relying on strong assumptions. The proposed method consistently outperforms state-of-the-art methods on the PACS and DomainNet datasets across various label fractions, showing significant improvements. Our code will be released. Our public code is available at \url{https://github.com/jinsuby/DomCLP}.

\end{abstract}

%% file: CameraReady/LaTeX/sections/1_introduction.tex
\section{Introduction}



Self-supervised learning (SSL) methods, particularly based on the instance discrimination task using contrastive learning with InfoNCE, have shown remarkable performance~\cite{infonce2,simclr,moco,byol,swav}.
Despite these successes, there is a critical limitation in that they assume pretraining, fine-tuning, and testing data all originate from the same distribution.
This assumption often does not hold in real-world scenarios, where data distribution shifts frequently occur, leading to previously unseen data.
Consequently, while SSL models are capable of generating high-quality representations for in-domain data, they often struggle to generate effective representations for unseen-domain data~\cite{darling,brad}. 
To address this challenge, there has been growing interest in the field of unsupervised domain generalization (UDG), which aims to develop SSL models that can learn domain-irrelevant features~\cite{darling}.

\begin{figure}[t]
\centering
\includegraphics[width=1.0\linewidth]{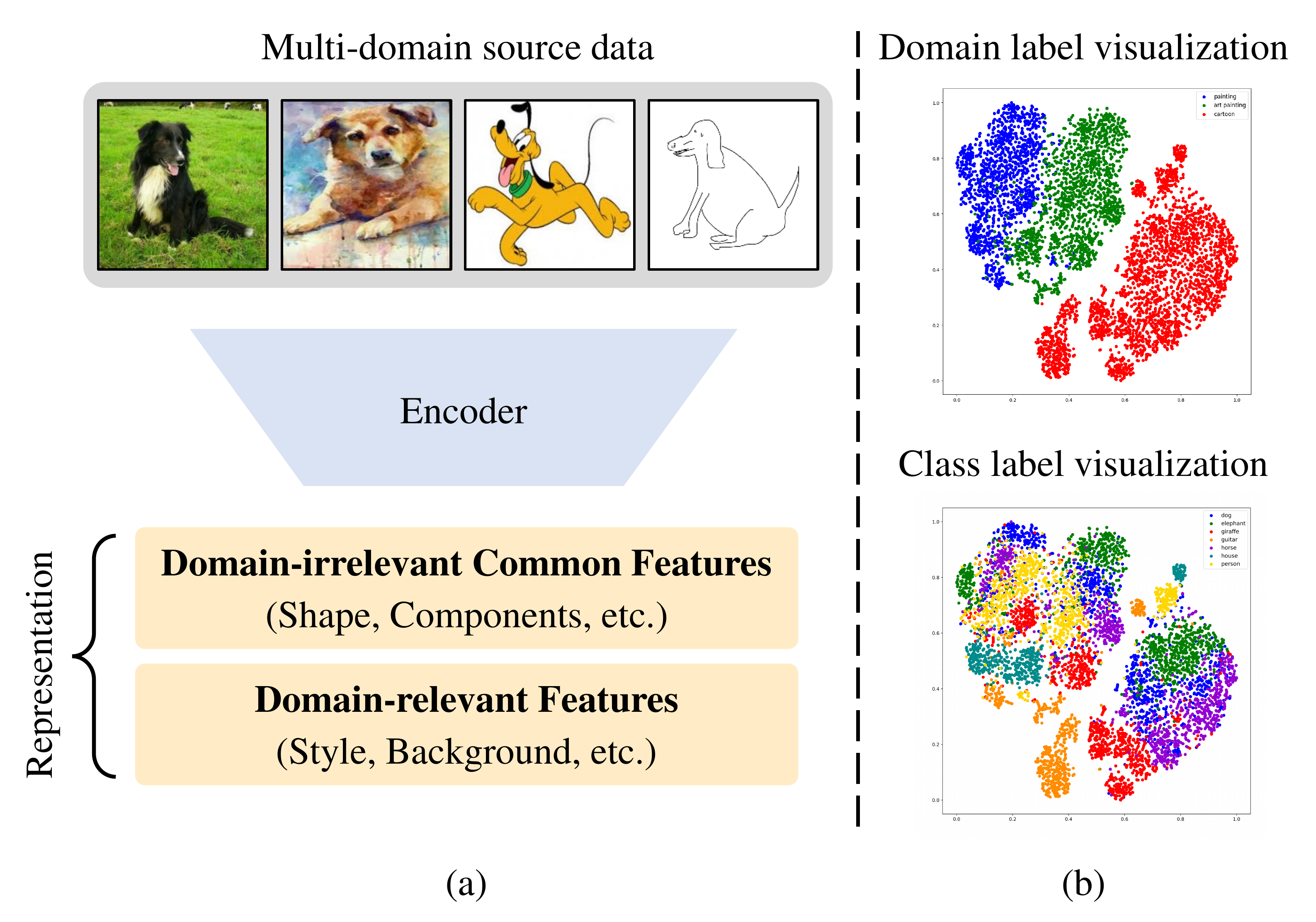}
\caption{(a) In the UDG environment, representations include both domain-irrelevant common features and domain-relevant features. (b) T-SNE visualization for SimCLR.}
\vspace{-5mm}
\label{figure1}
\end{figure}

In the UDG scenario as depicted in Figure 1a, there are multi-domain source datasets with only domain labels and no additional class labels.
Each sample contains both domain-irrelevant common features (e.g. shape, components, etc.) and domain-relevant features (e.g. style, background, etc.). 
To achieve good generalization performance on unseen domains, unsupervised domain generalization (UDG)~\cite{darling} aims to learn representations of domain-irrelevant feature.
To learn domain-irrelevant features, most UDG approaches utilize contrastive learning based on InfoNCE to generate feature representations, and perform feature alignment based on strong assumptions to achieve domain generalization~\cite{darling,brad,dn2a,bss}.
To align cross-domain features, BrAD~\cite{brad} utilized edge-like image transforms, while DN$^2$A~\cite{dn2a} employed cross-domain nearest neighbors as positive samples in contrastive learning.
In BSS~\cite{bss}, they introduced batch style standardization using Fourier transform for feature alignment. 
However, most approaches still struggle to effectively extract domain-irrelevant common features.

Most UDG approaches rely on instance discrimination tasks, which are not well-suited for domain generalization.
In contrastive learning with InfoNCE, representations are learned to distinguish between instances. 
If the model attempts to differentiate between instances, deep neural networks tend to learn features that are useful to discriminate instances. 
In UDG environments, they are easy to capture domain-relevant features rather than domain-irrelevant common features, because domain-relevant features are more helpful to distinguish instances across various domains~\cite{CanCLAvoidShortcut,chen2021intriguing, shortcut}. 
As a result, the instance discrimination task suppresses domain-irrelevant features and amplifies domain-relevant features, thereby hindering domain generalization, as shown in Figure 1b.

Another limitation is that previous approaches heavily depend on strong assumptions to align features across multi-domain.
\citet{brad} and \citet{dn2a} assumed that if two edge-like images or two images from different domains are similar, then the images will share domain-irrelevant common features.
\citet{bss} assumed that domain-irrelevant common features will not be lost or distorted even if an image is style transformed using a Fourier transform. 
To generalize domain-irrelevant common features, they tried to align features across multiple domains based on these strong assumptions.
However, these strong assumptions can lead to learning biased features or ignoring other important features. They reduce the diversity of domain-irrelevant common features, and only a limited set of common features is extracted~\cite{CanCLAvoidShortcut, attentionGeneralization, dn2a}.
To achieve high performance on unseen domains, we need to effectively generalize diverse domain-irrelevant common features, rather than relying on strong assumption based feature alignments that may lead to unintended bias.

To address these limitations, we propose a novel approach, \textbf{DomCLP}, \textbf{Dom}ain-wise \textbf{C}ontrastive \textbf{L}earning with \textbf{P}rototype Mixup for unsupervised domain generalization. 
First, we theoretically and experimentally demonstrate that some negative terms in InfoNCE can suppress domain-irrelevant common features and amplifies domain-relevant features. Building on this insight, we introduce the \underline{D}omain-wise \underline{Con}trastive Learning (DCon) to enhance domain-irrelevant common features while representation learning. 
Second, to effectively generalize diverse domain-irrelevant common features across multi-domain, we propose the \underline{P}rototype \underline{Mix}up Learning (PMix).
In PMix, to generalize common features from multi-domain, we interpolate common features in each domain utilizing mixup~\cite{mixup}. 
We extract prototypes of features by k-means clustering, and train the model with mixed prototypes by mixup.
It allows the model to effectively learn feature representations for unseen inter-manifold spaces while retaining diverse common feature information. Through our proposed method, DomCLP, the model effectively enhances and generalizes diverse common features.
We validate our approach through experiments on PACS and DomainNet datasets, achieving state-of-the-art performance with improvements of up to 11.3\% on the PACS 1\% label fraction and 12.32\% on the DomainNet 1\% label fraction.

%% file: CameraReady/LaTeX/sections/2_related.tex
\section{Related Works}
\subsection{Self-supervised Learning}

Self-supervised learning (SSL) aims to learn semantic features without relying on label information. 
Recently, most SSL methods have been based on contrastive learning with the information noise-contrastive estimation (InfoNCE) objective~\cite{infonce1,infonce2}, which has shown outstanding performance~\cite{simclr,simclrv2, moco,mocov2, byol,swav,adco,pcl}. 
These methods train models to bring augmented views of the same image closer together and pushing augmented views from different images farther apart. 
Although SSL models are effective at generating high-quality representations for samples within the same domain, they often struggle with unseen-domain data distributions.
In real-world scenarios, data distribution shifts often occur, causing self-supervised models to generate less effective representations and leading to degraded generalization performance.
To address this challenge, unsupervised domain generalization (UDG) has been proposed~\cite{darling}.
\subsection{Unsupervised Domain Generalization}
Self-supervised learning aims to generalize well on the given training data, while unsupervised domain generalization (UDG) aims to generalize well on unseen domain data.
To extract good representations for unseen domains, the model needs to learn domain-irrelevant common features.
To achieve this, most UDG approaches utilize contrastive learning with InfoNCE and strong assumptionss to align features.
DARLING~\cite{darling} first proposed the UDG task and introduced a new learning technique based on a graphical probability model. BrAD~\cite{brad} proposed a self-supervised cross-domain learning method that semantically aligns all domains to an edge-like domain, while DN2A~\cite{dn2a} utilized strong augmentations to destroy intra-domain connectivity and dual nearest neighbors to align cross-domain features.
BSS~\cite{bss} introduced batch styles standardization using Fourier transforms to align features with transformed images.
These methods have improved performance in UDG, but since most UDG approaches utilize contrastive learning with InfoNCE, they often suppress domain-irrelevant common features while amplifying domain-relevant features.
Additionally, the strong assumptions used in existing approaches can lead to learning biased features or ignoring other important features, which reduces the diversity of domain-irrelevant common features~\cite{shortcut, li2020shape, CanCLAvoidShortcut, chen2021intriguing}.

%% file: CameraReady/LaTeX/sections/3_proposed.tex
\section{Proposed Method}

We aim to effectively enhance domain-irrelevant common features and generalize the common features across multi-domain.
To achieve this goal, we propose a novel approach, DomCLP, \textbf{Dom}ain-wise \textbf{C}ontrastive \textbf{L}earning with \textbf{P}rototype Mixup for unsupervised domain generalization.

In this section, we show that InfoNCE is not effective to extract domain-irrelevant features. We theoretically explore how  InfoNCE leads to suppress domain-irrelevant common features and amplify domain-relevant features. Based on this analysis, we propose  the \underline{D}omain-wise \underline{Con}trastive Learning (DCon) to generate feature representations with enhanced domain-irrelevant common features.
Furthermore, we present the \underline{P}rototype \underline{Mix}up Learning (PMix) to generalize diverse common features across multi-domain. 
While strong assumption-based feature alignment methods reduce feature diversity during generalization, our method, PMix, effectively generalizes feature representations to unseen domains while maintaining diverse common features from multiple domains through mixup.

\subsection{Problem Formulation of UDG}
In the UDG setting, multi-domain source datasets $S = \{(x_i, y^d_i)\}_{i=1}^{N_S}$ are provided, containing domain labels $y^d$ but no class labels $y^c$. The model is trained to generate domain-irrelevant common feature representations from these datasets. To evaluate the encoder's ability to extract common features, a classifier is trained on a subset of the multi-domain source labeled data $S_L = \{(x_i, y^c_i, y^d_i)\}_{i=1}^{N_{S_L}}$ while keeping the encoder frozen. The model's performance is  assessed using an unseen-domain target dataset $T = \{(x_i, y^c_i, y^d_i)\}_{i=1}^{N_T}$ to verify how effectively the trained encoder extracts high-quality common features.

\subsection{DCon: \underline{D}omain-wise \underline{Con}trastive Learning}
In contrastive-based SSL methods, augmented samples from the original image of an anchor sample are treated as positive samples and pulled closer, while other samples in the batch are treated as negative pairs and pushed farther apart. The InfoNCE loss is defined as follows:
\begin{equation}
\mathcal{L}_\text{Info}^{i} = -\log \frac{\exp(z_i \cdot z_i^+ /\tau)}{\sum_{k=1}^{2N} \mathds{1}_{[k \neq i]} \exp(z_i \cdot z_k^-/\tau)}
\label{equation1}
\end{equation}
\noindent The representations of the anchor, positive, and negative samples are denoted as $z_i$, $z_i^+$, and $z^-$, respectively. Additionally, $N$ refers to the batch size, and $\tau$ represents the temperature.
In contrastive learning with InfoNCE, representations are learned to distinguish between instances. If multi-domain datasets are given, it is more likely to capture domain-relevant features that are easier to distinguish, rather than domain-irrelevant common features that are harder to distinguish~\citep{CanCLAvoidShortcut, chen2021intriguing}. 
As a result, instance discrimination tasks using InfoNCE tend to suppress domain-irrelevant features and amplify domain-relevant features.

To verify that InfoNCE is not well-suited for learning domain-irrelevant common features, we have reformulated \Cref{equation1} as follows:
%
%
%
\begin{align}
\mathcal{L}_\text{Info}^{i} &= -\log \frac{\exp(z_i \cdot z_i^+ /\tau)}{\exp(z_i \cdot z_i^+ /\tau)+N^{\alpha}_i + N^{\beta}_i + N^{\gamma}_i} \\
N^{\alpha}_{i} &= \sum_{k} 
\mathds{1}_{[k \neq i, k \in D_i]}
\exp(z_i \cdot z_k^-/\tau) \nonumber \\
N^{\beta}_{i} &= \sum_{k} 
\mathds{1}_{[k\neq i, k \in D^C_i \cap F_i]}
\exp(z_i \cdot z_k^-/\tau) \nonumber \\
N^{\gamma}_{i} &= \sum_{k} \mathds{1}_{[k \neq i, D^C_i \cap F^C_i]} 
\exp(z_i \cdot z_k^-/\tau) \nonumber \\
D_i &= \{ j\:|\:y^d_j = y^d_i \} \nonumber \\
F_i &= \{ j\:|\: \epsilon > |\delta_{ij}|, \delta_{ij}= c_j - c_i \} \nonumber
\label{equation2}
\end{align}

\begin{figure}[t]
\centering
\includegraphics[width=0.5\linewidth]{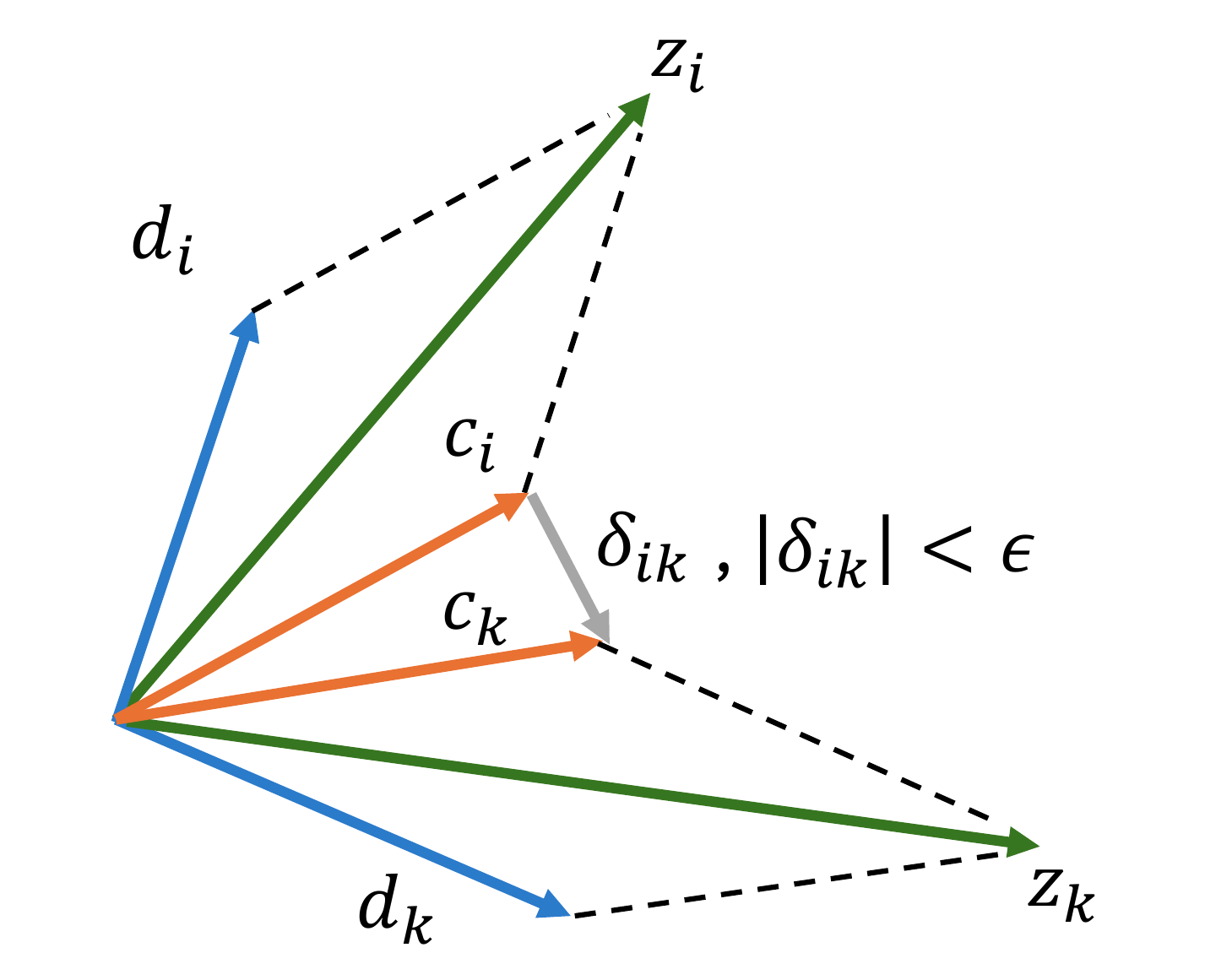}

\caption{Each representation $z_i$ consists of domain-irrelevant common features $c_i$ and domain-relevant features.}
\label{figure2}
\vspace{-5mm}
\end{figure}

\noindent
We assume that the representation $z_i$ can be expressed as $d_i + c_i$ (depicted in \Cref{figure2}), where $d_i$ and $c_i$ represent domain-relevant and domain-irrelevant common features, respectively. 
$D_i$ is the set of samples from the same domain of $x_i$, and $F_i$ is the set of samples whose common features are within a small distance $\epsilon$ of those of $x_i$.
The complement of $A$ is denoted by $A^C$. 
Then, $N^\alpha_i$, $N^\beta_i$, and $N^\gamma_i$ denote the negative pairs in the same domain, the negative pairs having similar common features in the different domains, and the negative pairs with both different domains and different common features, respectively. Since $c_k=c_i+\delta_{ik}$, $N^\beta_i$ can be rewritten as follows:
\begin{equation}
N^{\beta}_{i} = \sum_{k\neq i, k \in D^C_i \cup F_i} 
\exp((c_i + d_i) \cdot (c_i + d_k + \delta_{ik})/\tau)
\label{equation3}
\end{equation}
To minimize $\mathcal{L}_\text{Info}^{i}$, the negative terms $N^{\beta}_{i}$ must be minimized. It is clear that $|c_i|$ and $d_i \cdot d_k$ should be minimized. Since $x_i$ and $x_k$ are in different domains, we may assume that $d_i$ and $d_k$ are not parallel or antiparallel to each other. 
Since $\delta_{ik}$ is negligible, and $x_i$ and $x_k$ are neither parallel nor antiparallel, $|c_i|$ will be close to 0, and $d_i \cdot d_k$ will be a large negative value. 
In other words, the common features, $c_i$, will be suppressed, while the domain-relevant features, $d_i$ and $d_k$, will be amplified.

To prevent these problems, $N^{\beta}_{i}$ should not be included in the negative terms. However, in a UDG environment, it is difficult to accurately identify $F_{i}$ and $F^C_{i}$ for given $x_i$ because we cannot decompose $z_i$ into $c_i$ and $d_i$.
Therefore, we exclude both $N^{\beta}_{i}$ and $N^{\gamma}_{i}$ from the negative terms. 
In other words, our method performs domain-wise contrastive learning (DCon), and the objective is as follows:
\begin{equation}
\mathcal{L}_\text{dcon} = \sum_{i}-\log \frac{\exp(z_i \cdot z_i^+ /\tau)}{\sum_{k} \mathds{1}_{[k \neq i, k \in D_i]} \exp(z_i \cdot z_k/\tau)}
\label{equation4}
\end{equation}




\begin{figure}[t]
\centering
\includegraphics[width=1.0\linewidth]{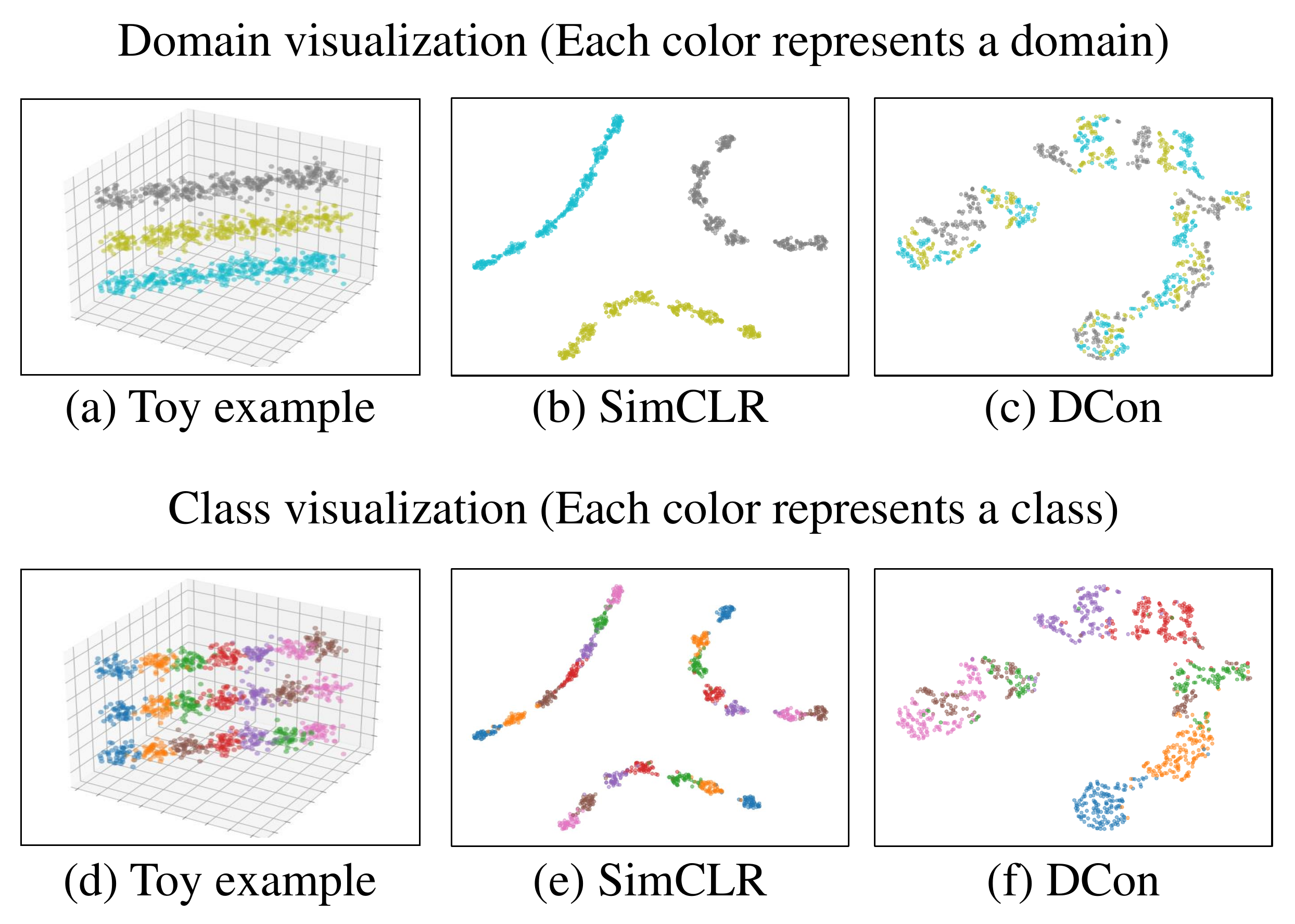}
\caption{T-SNE visualization of a multi-domain 3D toy example. (a) and (d) are the toy example colored by domain and class, respectively. (b) and (e) are t-SNE visualizations for SimCLR. (c) and (f) are t-SNE visualizations for DCon.}
\vspace{-2mm}
\label{figure3}
\vspace{-1mm}
\end{figure}


\Cref{figure3} demonstrates the effectiveness of DCon. We present the t-SNE~\cite{tsne} results of features by DCon and SimCLR on a 3D toy example.
The figures in the upper row and lower row are colored by domain labels and class labels, respectively.
Even though most classes in the toy example are mostly aligned across domains, the representations from SimCLR are mainly clustered based on domains rather than classes because the model captures domain-relevant features instead of domain-irrelevant ones. In contrast, the representations from DCon are mainly clustered according to classes, as shown in Figure 3c and 3f. 
This example shows how our DCon is effective to extract domain-irrelevant common features.

\subsection{PMix: \underline{P}rototype \underline{Mix}up Learning}
Existing UDG approaches tried to generalize common features from multiple domains based on feature alignment. They merge each domain’s feature manifold into a single manifold through strong assumption-based feature alignments~\cite{brad,bss,dn2a}. However, these strong assumption-based feature alignments can lead to learning biased features or reducing the diversity of common features~\cite{shortcut, li2020shape, CanCLAvoidShortcut, chen2021intriguing}.

\begin{figure}[t]
\centering
\includegraphics[height=4.4cm]{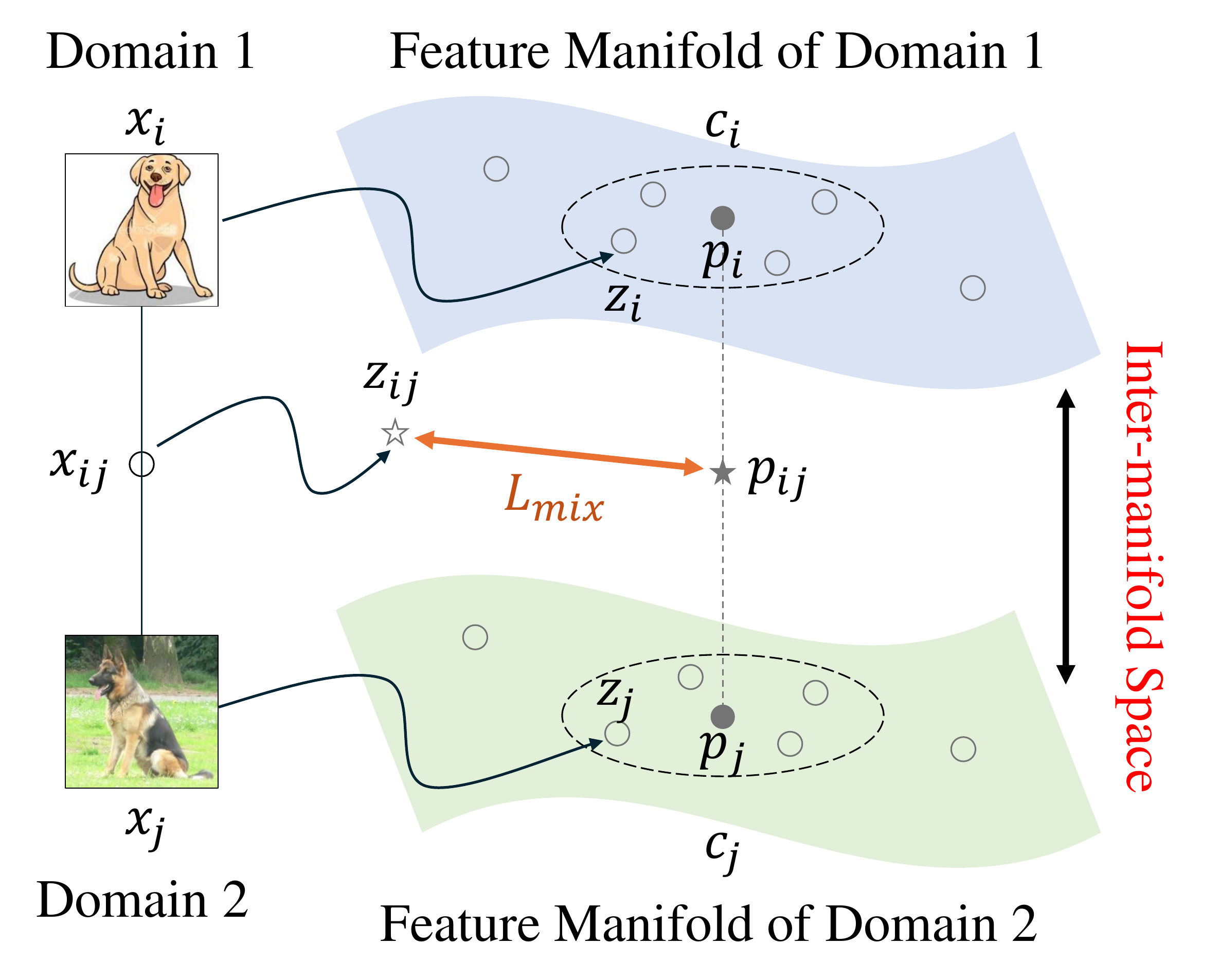}

\caption{The framework for Prototype Mixup Learning.}
\vspace{-3mm}
\label{figure4}
\vspace{-1mm}
\end{figure}

We do not try to merge each domain’s feature manifold into a single manifold. We facilitate representation learning for the inter-manifold space using mixup~\cite{mixup}. For example, \Cref{figure4} shows the representations of samples from two domains. Each sample places on its respective manifold. 
Since the model has learned only how to map samples of seen source domains onto the appropriate manifolds, it may fail to generate proper representations for unseen domain data. For instance, let $z_i$ be the representation of an image $x_i$ from Domain 1, and $z_j$ be the representation of an image $x_j$ from Domain 2. Then, the mixup of $x_i$ and $x_j$, denoted as $x_{ij}$, can be regarded as a sample from an unseen domain by the model. The model would generate an improper representation $z_{ij}$ for $x_{ij}$. 

Since our goal is to train the model to extract domain-irrelevant common features from any sample, we train the model to map $x_{ij}$ to a mixup of the common features of $x_i$ and $x_j$. This will encourage the model to learn how to map samples from an unseen domain into a new manifold within the representation space. 
However, since we cannot separate domain-irrelevant common features from $z_i$ and $z_j$, we approximate them through clustering.
We perform k-means clustering on each domain separately, and find the clusters $c_i$ and $c_j$ which contain $x_i$ and $x_j$, respectively. We use the prototypes, which are the centroids of clusters, $p_i$ and $p_j$ of each cluster as the common features of $x_i$ and $x_j$. Since samples with similar representations are grouped into a cluster, their average can be considered as their common feature. 

To generalize common features for unseen domain, we need a diverse set of common features~\cite{zhang2022use, jiang2023hierarchical}.
To learn common features from diverse perspectives, we use multiple clustering results. If we want to extract fine-grained or specific common features, we may cluster with a large number of clusters. Conversely, if we need more general or broader common features, we can cluster with a small number of clusters.
We perform k-means clustering with multiple
numbers of clusters, $K = \{k_1, \cdots , k_M \}$. For a cluster number, $k_m$, we cluster each domain separately into $k_m$ clusters, and combine the whole clusters from each domain, which is denoted by $C^m$.
We find the cluster in $C^m$ to which $z_i$ belongs, and denote its centroid as $p^m_i$.
The loss function for PMix is as follows:
\begin{equation}
\begin{split}
\mathcal{L}_\text{pmix} &= \frac{1}{NM} \sum_{i=1}^N \sum_{m=1}^M \|p_{ij}^m - f(\theta(x_{ij})) \|_2
\\
x_{ij} &= \lambda x_i + (1 - \lambda) x_j
\\
p_{ij}^m &= \lambda p_i^m + (1 - \lambda) p_j^m
\end{split}
\label{equation5}
\end{equation}
\noindent where $\lambda \sim \text{Beta}(\alpha, \alpha)$ is a mixing coefficient sampled from the Beta distribution~\cite{mixup}, and $x_j$ is a randomly selected from the batch. The representations are extracted through the encoder $\theta$ and the projection head $f$. $M$ is the number of clustering results, and $N$ is the number of training samples.
We also apply this mixup-based interpolation to samples within the same domain, as it also helps learning the feature representations within the same domain~\cite{mixup,lsl}.

Additionally, to enhance clustering quality and extract well-representative prototypes, we employ prototypical contrastive learning~\cite{pcl}. It makes each representation closer to its corresponding prototype, and farther from other prototypes. The loss function for prototypical contrastive learning is as follows:
\begin{equation}
\mathcal{L}_\text{pcl} = \frac{1}{NM} \sum_{i=1}^N \sum_{m=1}^{M} -\log \frac{\exp(z_i \cdot p_i^m / \phi_i^m)}{\sum_{j=1}^{k_m} \exp(z_i \cdot p_j^m/\phi_j^m)}
\label{equation6}
\end{equation}
where $\phi$ denotes the concentration estimation, where a smaller $\phi$ indicates larger concentration. 
The overall objectives can be summarized as follows:
\begin{equation}
\mathcal{L} = \mathcal{L}_\text{dcon} + \mathcal{L}_\text{pmix} + \mathcal{L}_\text{pcl}
\label{equation7}
\end{equation}

The overall algorithm related to the total objectives is provided in the supplementary material. With our proposed objectives, the model effectively enhances and generalizes domain-irrelevant common features without relying on strong assumptions.



%% file: CameraReady/LaTeX/sections/4_experiments.tex
\section{Experiments} 

\begin{table}[!t]
\centering
\footnotesize
\setlength{\tabcolsep}{0.1cm}          
\begin{tabular}{lcccc|c}
\toprule 
Target domain & Photo & Art. & Cartoon & Sketch & Avg. 
\\ \midrule

\multicolumn{6}{c}{Label Fraction 1\%}
\\ \midrule 

ERM & 10.90 & 11.21 & 14.33 & 18.83 & 13.82
\\ 

MoCo V2 & 22.97 & 15.58 & 23.65 & 25.27 & 21.87
\\ 

BYOL & 11.20 & 14.53 & 16.21 & 10.01 & 12.99
\\ 

AdCo & 26.13 & 17.11 & 22.96 & 23.37 & 22.39
\\ 

SimCLR V2 & 30.94 & 17.43 & 30.16 & 25.20 & 25.93
\\ 

DARLING & 27.78 & 19.82 & 27.51 & 29.54 & 26.16
\\ 

BrAD (kNN) & 55.00 & 35.54 & 38.12 & 34.14 & 40.70
\\ 

BrAD (linear) & 61.81 & 33.57 & 43.47 & 36.37 & 43.81
\\ 

DN$^2$A (kNN) & 66.37 & 42.68 & 49.85 & 54.37 & 53.32 \\ 

DN$^2$A (linear) & 69.15 & 46.04 & 51.19 & 56.88 & 55.82
\\ 

SimCLR w/ BSS (linear) & 43.31 & 38.96 & 48.61 & 48.76 & 44.91
\\ 

SWaV w/ BSS (linear) & 39.74 & 35.82 & 42.59 & 36.12 & 38.57
\\ \midrule 

Ours (KNN) & 66.51 & 56.43 & 53.29 & 61.79 & 59.50
\\

Ours (linear) & \textbf{71.04} & \textbf{59.20} & \textbf{56.10} & \textbf{62.15} & \textbf{62.13} 
\\ \midrule

\multicolumn{6}{c}{Label Fraction 5\%}
\\ \midrule 

ERM & 14.15 & 18.67 & 13.37 & 18.34 & 16.13
\\ 

MoCo V2 & 37.39 & 25.57 & 28.11 & 31.16 & 30.56
\\ 

BYOL & 26.55 & 17.79 & 21.87 & 19.65 & 21.46
\\ 

AdCo & 37.65 & 28.21 & 28.52 & 30.56 & 31.18
\\ 

SimCLR V2 & 54.67 & 35.92 & 35.31 & 36.84 & 40.68
\\ 

DARLING & 44.61 & 39.25 & 36.41 & 36.53 & 39.20
\\ 

BrAD (kNN) & 58.66 & 39.11 & 45.37 & 46.11 & 47.31
\\ 

BrAD (linear) & 65.22 & 41.35 & 50.88 & 50.68 & 52.03
\\ 

DN$^2$A (kNN) & 68.93 & 46.83 & 54.40 & 59.92 & 57.52
\\ 

DN$^2$A (linear) & 73.16 & 52.20 & 59.75 & 66.43 & 62.89
\\ 

SimCLR w/ BSS (linear) & 58.16 & 46.37 & 55.69 & 65.63 & 56.40
\\ 

SWaV w/ BSS (linear) & 50.58 & 43.00 & 53.81 & 52.61 & 50.00
\\ \midrule 

Ours (kNN) & 70.74 & 60.71 & 57.01 & 67.34 & 63.95 
\\ 

Ours (linear) & \textbf{76.79} & \textbf{61.81} & \textbf{62.50} & \textbf{71.87} & \textbf{68.24}
\\ \midrule

\multicolumn{6}{c}{Label Fraction 10\%}
\\ \midrule 

ERM & 16.27 & 16.62 & 18.40 & 12.01 & 15.82
\\ 

MoCo V2 & 44.19 & 25.85 & 35.53 & 24.97 & 32.64
\\ 

BYOL & 27.01 & 25.94 & 20.98 & 19.69 & 23.40
\\ 

AdCo & 46.51 & 30.31 & 31.45 & 22.96 & 32.81
\\ 

SimCLR V2 & 54.65 & 37.65 & 46.00 & 28.25 & 41.64
\\ 

DARLING & 53.37 & 39.91 & 46.41 & 30.17 & 42.46 
\\ 

BrAD (kNN) & 67.20 & 41.99 & 45.32 & 50.04 & 51.14
\\ 

BrAD (linear) & 72.17 & 44.20 & 50.01 & 55.66 & 55.51
\\ 

DN$^2$A (kNN) & 69.73 & 50.29 & 59.22 & 64.95 & 61.05
\\ 

DN$^2$A (linear) & 75.41 & 53.14 & 63.69 & 68.57 & 65.20
\\ 

SimCLR w/ BSS (linear) & 63.29 & 51.37 & 59.43 & 66.09 & 60.04
\\ 

SWaV w/ BSS (linear) & 57.82 & 45.91 & 53.65 & 55.67 & 53.27
\\ \midrule 

Ours (kNN) & 70.32 & 59.86 & 59.44 & 67.76 & 64.35 
\\ 

Ours (linear) & \textbf{77.08} & \textbf{65.45} & \textbf{63.99} & \textbf{73.44} & \textbf{69.99}
\\  \bottomrule

\end{tabular}

\caption{UDG performances on PACS dataset. To evaluate the target domain, linear and kNN (non-parametric) classifiers are trained on a few labeled samples from the three source domains. ERM is the randomly initialized model. Most of the experimental results are extracted from state-of-the-art methods~\cite{darling, brad, dn2a, bss}.
\textbf{Bold values} indicate best performances, and all experiments are conducted for 3 folds.}
\label{table1}
\end{table}

\begin{table*}[t]
\centering
\aboverulesep=0.2ex 
\belowrulesep=0.2ex 
\scriptsize
\setlength{\tabcolsep}{0.5cm}          
\renewcommand{\arraystretch}{0.2}       
\begin{tabular}{lcccccc|cc}
\toprule\rule{0pt}{1.0EM}

Source domains & \multicolumn{3}{c}{\{Paint $\cup$ Real $\cup$ Sketch\}} & \multicolumn{3}{c}{\{Clipart $\cup$ Info. $\cup$ Quick.\}} & & 
\\ \rule{0pt}{1.0EM}

Target domains & Clipart & Info. & Quick. & Painting & Real & Sketch & Overall & Avg.
\\ \midrule

\multicolumn{9}{c}{\rule{0pt}{1.0EM}Label Fraction 1\%}
\\ \midrule\rule{0pt}{1.0EM}

ERM & 6.54 & 2.96 & 5.00 & 6.68 & 6.97 & 7.25 & 5.88 & 5.89
\\ \rule{0pt}{1.0EM}

BYOL & 6.21 & 3.48 & 4.27 & 5.00 & 8.47 & 4.42 & 5.61 & 5.31
\\ \rule{0pt}{1.0EM}

MoCo V2 & 18.85 & 10.57 & 6.32 & 11.38 & 14.97 & 15.28 & 12.12 & 12.90
\\ \rule{0pt}{1.0EM}

AdCo & 16.16 & 12.26 & 5.65 & 11.13 & 16.53 & 17.19 & 12.47 & 13.15
\\ \rule{0pt}{1.0EM}

SimCLR V2 & 23.51 & 15.42 & 5.29 & 20.25 & 17.84 & 18.85 & 15.46 & 16.55
\\ \rule{0pt}{1.0EM}

DARLING & 18.53 & 10.62 & 12.65 & 14.45 & 21.68 & 21.30 & 16.56 & 16.53
\\ \rule{0pt}{1.0EM}


BrAD (kNN) & 40.65 & 14.00 & 21.28 & 16.80 & 22.29 & 25.72 & 22.35 & 23.46
\\ \rule{0pt}{1.0EM}

BrAD (linear) & 47.26 & 16.89 & 23.74 & 20.03 & 25.08 & 31.67 & 25.85 & 27.45
\\ \rule{0pt}{1.0EM}


DN$^2$A (kNN) & 62.31 & 23.84 & 27.50 & 29.71 & 37.07 & 45.48 & 35.21 & 37.65
\\ \rule{0pt}{1.0EM}

DN$^2$A (linear) & 68.02 & 24.45 & 29.20 & 31.16 & 37.91 & 52.62 & 37.43 & 40.56
\\ \rule{0pt}{1.0EM}

SimCLR w/ BSS (linear) & 61.94 & 19.58 & 26.98 & 27.40 & 31.55 & 41.49 & 32.27 & 34.82
\\ \rule{0pt}{1.0EM}

SWaV w/ BSS (linear) & 60.40 & 20.12 & 23.09 & 34.64 & 38.45 & 46.90 & 34.32 & 37.27
\\ \midrule\rule{0pt}{1.0EM}

Ours (kNN) & 66.06 & 24.93 & 31.25 & 31.77 & 38.14 & 51.53 & 37.89 & 40.61 
\\ \rule{0pt}{1.0EM}

Ours (linear) & \textbf{70.31} & \textbf{28.49} & \textbf{38.10} & \textbf{38.37} & \textbf{43.29} & \textbf{54.81} & \textbf{43.18} & \textbf{45.56}
\\ \midrule

\multicolumn{9}{c}{\rule{0pt}{1.0EM}Label Fraction 5\%}
\\ \midrule\rule{0pt}{1.0EM}

ERM & 10.21 & 7.08 & 5.34 & 7.45 & 6.08 & 5.00 & 6.50 & 6.86
\\ \rule{0pt}{1.0EM}

BYOL & 9.60 & 5.09 & 6.02 & 9.78 & 10.73 & 3.97 & 7.83 & 7.53
\\ \rule{0pt}{1.0EM}

MoCo V2 & 28.13 & 13.79 & 9.67 & 20.80 & 24.91 & 21.44 & 18.99 & 19.79 
\\ \rule{0pt}{1.0EM}

AdCo & 30.77 & 18.65 & 7.75 & 19.97 & 24.31 & 24.19 & 19.42 & 20.94
\\ \rule{0pt}{1.0EM}

SimCLR V2 & 34.03 & 17.17 & 10.88 & 21.35 & 24.34 & 27.46 & 20.89 & 22.54
\\ \rule{0pt}{1.0EM}

DARLING & 39.32 & 19.09 & 10.50 & 21.09 & 30.51 & 28.49 & 23.31 & 24.83
\\ \rule{0pt}{1.0EM}

BrAD (kNN) & 55.75 & 18.15 & 26.93 & 24.29 & 33.33 & 37.54 & 31.12 & 32.66
\\ \rule{0pt}{1.0EM}

BrAD (linear) & 64.01 & 25.02 & 29.64 & 29.32 & 34.95 & 44.09 & 35.37 & 37.84 
\\ \rule{0pt}{1.0EM}

DN$^2$A (kNN) & 66.54 & 23.98 & 34.47 & 37.89 & 44.65 & 54.57 & 41.64 & 43.68
\\ \rule{0pt}{1.0EM}

DN$^2$A (linear) & 70.10 & \textbf{27.31} & 36.77 & 40.93 & 47.20 & 60.05 & 44.98 & 47.06
\\ \rule{0pt}{1.0EM}

SimCLR w/ BSS (linear) & 71.21 & 20.93 & 32.42 & 36.68 & 41.49 & 52.75 & 39.73 & 42.58 
\\ \rule{0pt}{1.0EM}

SWaV w/ BSS (linear) & 70.56 & 24.35 & 28.83 & \textbf{46.17} & 51.21 & 59.71 & 43.53 & 46.81
\\ \midrule\rule{0pt}{1.0EM}

Ours (kNN) & 68.52 & 26.23 & 35.59 & 39.61 & 48.11 & 58.20 & 43.94 & 46.04
\\ \rule{0pt}{1.0EM}

Ours (linear) & \textbf{73.44} & 25.18 & \textbf{38.81} & 43.40 & \textbf{51.38} & \textbf{62.02} & \textbf{46.92} & \textbf{49.04}
\\ \midrule

\multicolumn{9}{c}{\rule{0pt}{1.0EM}Label Fraction 10\%}
\\ \midrule\rule{0pt}{1.0EM}

ERM & 15.10 & 9.39 & 7.11 & 9.90 & 9.19 & 5.12 & 8.94 & 9.30
\\ \rule{0pt}{1.0EM}

BYOL & 14.55 & 8.71 & 5.95 & 9.50 & 10.38 & 4.45 & 8.69 & 8.92
\\ \rule{0pt}{1.0EM}

MoCo V2 & 32.46 & 18.54 & 8.05 & 25.35 & 29.91 & 23.71 & 21.87 & 23.05
\\ \rule{0pt}{1.0EM}

AdCo & 32.25 & 17.96 & 11.56 & 23.35 & 29.98 & 27.57 & 22.79 & 23.78
\\ \rule{0pt}{1.0EM}

SimCLR V2 & 37.11 & 19.87 & 12.33 & 24.01 & 30.17 & 31.58 & 24.28 & 25.84
\\ \rule{0pt}{1.0EM}

DARLING & 35.15 & 20.88 & 15.69 & 25.90 & 33.29 & 30.77 & 26.09 & 26.95
\\ \rule{0pt}{1.0EM}

BrAD (kNN) & 60.78 & 19.76 & 31.56 & 26.06 & 37.43 & 41.38 & 34.77 & 36.16
\\ \rule{0pt}{1.0EM}

BrAD (linear) & 68.27 & 26.60 & 34.03 & 31.08 & 38.48 & 48.17 & 38.74 & 41.10
\\ \rule{0pt}{1.0EM}

DN$^2$A (kNN) & 66.73 & 22.15 & 35.93 & 36.42 & 46.12 & 57.14 & 42.21 & 44.08
\\ \rule{0pt}{1.0EM}

DN$^2$A (linear) & 73.04 & \textbf{28.23} & 37.80 & 41.77 & 50.94 & 61.69 & 46.72 & 48.91
\\ \rule{0pt}{1.0EM}

SimCLR w/ BSS (linear) & 71.95 & 21.27 & 33.47 & 39.49 & 44.67 & 55.42 & 41.57 & 44.38
\\ \rule{0pt}{1.0EM}

SWaV w/ BSS (linear) & 71.99 & 24.34 & 29.82 & \textbf{48.28} & \textbf{52.37} & 60.55 & 44.59 & 47.89
\\ \midrule\rule{0pt}{1.0EM}

Ours (kNN) & 70.39 & 25.19 & 36.58 & 40.25 & 50.70 & 58.96 & 45.11 & 47.01
\\ \rule{0pt}{1.0EM}

Ours (linear) & \textbf{73.18} & 27.02 & \textbf{39.21} & 40.45 & 51.75 & \textbf{63.70} & \textbf{47.09} & \textbf{49.22} 
\\ \bottomrule

\end{tabular}

\caption{UDG performances on DomainNet dataset. \textbf{Bold values} indicate best performances.}
\label{table2}
\vspace{-2mm}
\end{table*}

\subsection{Setting and Datasets}
To verify the effectiveness of our method, we conduct experiments on commonly used UDG benchmark datasets, such as PACS~\cite{pacs} and DomainNet~\cite{domainnet}. 

PACS dataset~\cite{pacs} consists four distinct domains: Photo, Art painting, Cartoon, and Sketch. Each domain contains images from the same seven categories: dog, elephant, giraffe, guitar, house, horse, and person. In total, there are 9,991 images, with each image having dimensions of 224$\times$224$\times$3. This dataset is widely used to evaluate how well a model can generalize across different visual styles, from realistic photos to highly abstract sketches. For training, the other three domains, excluding the target domain, are used as source domains.

DomainNet dataset~\cite{domainnet} is a large-scale dataset for domain generalization. It consists of images from six distinct domains: Real, Clipart, Painting, Sketch, Infograph, and Quickdraw. The full dataset covers 345 categories, however, we use 20 sub-categories following to the existing UDG protocol~\cite{darling, brad, dn2a, bss}. Each image is of varying sizes, typically around 224$\times$224$\times$3. The diverse domains and large number of categories present a challenging scenario for models to learn domain-invariant features. For training, the other three domains, excluding the three target domains, are used as source domains.
Additional details are provided in supplementary material.

\subsection{Implementation Details}
All experiments are conducted using the PyTorch framework on an NVIDIA RTX 3090Ti GPU. 
To ensure a fair comparison, we evaluate our approach against existing SSL methods~\cite{byol, mocov2, adco, simclrv2} and UDG methods ~\cite{darling, brad, dn2a, bss}, and all conditions are set according to the protocols outlined in existing UDG methods.
For the PACS dataset, we use a non-pretrained ResNet-18~\cite{resnet}, Adam optimizer~\cite{adam} for 1000 epochs, a weight decay of 1e-4, an initial learning rate of 3e-4, and the cosine annealing function as a learning scheduler. 
We set numbers of clusters to $K = \{7,14,28\}$, $\tau_r$ to 0.07, and the batch size to 256.
For the DomainNet dataset, we use a pretrained ResNet-18, Adam optimizer for 1000 epochs, a weight decay of 1e-4, an initial learning rate of 3e-4, and the cosine annealing function as a learning scheduler. We set numbers of clusters $K = \{20,40,80\}$, $\tau_r$ to 0.07, and the batch size to 256.
To evaluate the target domain, linear and kNN (non-parametric) classifiers are trained on a few labeled samples from the three source domains.
For mixup interpolation, we set both $\alpha$ and $\beta$ to 4 for beta mixture on all experiments. 
According to \citet{dn2a}, strong augmentation is beneficial for UDG, so we use RandAugment~\cite{randaug} as the strong augmentation.

\subsection{Main Results}


\subsubsection{Results for PACS.}
The results for the PACS dataset under various label fractions are shown in \Cref{table1}. Our proposed method consistently outperforms other state-of-the-art methods across all target domains and label fractions. Notably, our method significantly surpasses others in the 1\%, 5\%, and 10\% label fractions, with average performance improvements of 11.3\%, 8.5\%, and 7.3\%, respectively. These results clearly indicate that our method is highly effective for UDG tasks.

\subsubsection{Results for DomainNet.}
\Cref{table2} shows the results for the DomainNet dataset under various label fractions. Our proposed method demonstrates superior performance across all target domains in the 1\% label fraction. In the 5\% and 10\% label fractions, it shows the best or highly competitive performance. Overall, our method consistently achieves higher average accuracy compared to existing methods across all label fractions, with a particularly notable average improvement of 12.32\% in the 1\% label fraction.


\subsection{Ablation Studies}
All experiments in ablation studies are conducted on PACS dataset using 5\% labeled data and kNN for 3 folds. 
\subsubsection{Effectiveness of components.}
\Cref{table3} presents the performance of our method on the PACS dataset with different combinations of our proposed modules. Using only $L_{dcon}$ achieves an average accuracy of 60.56\%. When utilizing both $L_{pmix}$ and $L_{pcl}$, the model's accuracy significantly increases to 63.95\%, compared to using each module individually.
This improvement is due to $L_{pmix}$ facilitating the generalization of diverse common features across multiple domains, while $L_{pcl}$ enhances the extraction of well-representative prototypes.

\begin{table}[t]
\centering
\aboverulesep=0ex 
\belowrulesep=0ex 
\small
\setlength{\tabcolsep}{0.1cm}          
\begin{tabular}{ccc|cccc|c}
\toprule \rule{0pt}{1.0EM}
$L_{dcon}$ & $L_{pmix}$ & $L_{pcl}$ & Photo & Art. & Cartoon & Sketch & Avg.
\\ \midrule \rule{0pt}{1.0EM}

\cmark & \xmark & \xmark & 72.40 & 53.79 & 51.09 & 64.95 & 60.56
\\ \rule{0pt}{1.0EM}

\cmark & \cmark & \xmark & \textbf{72.18} & 58.66 & 51.38 & 64.24 & 61.62 
\\ \rule{0pt}{1.0EM}

\cmark & \xmark & \cmark & 69.92 & 56.28 & 52.23 & 65.96 & 61.10 
\\ \midrule \rule{0pt}{1.0EM}

\cmark & \cmark & \cmark & 70.74 & \textbf{60.71} & \textbf{57.01} & \textbf{67.34} & \textbf{63.95} \\ \bottomrule
\end{tabular}

\caption{Performance on PACS with our proposed modules. $L_{dcon}$, $L_{pmix}$, and $L_{pcl}$ represent the loss for domain-wise contrastive learning, prototype mixup, and prototypical contrastive learning, respectively.}
\label{table3}
\end{table}

\begin{table}[t]
\centering
\aboverulesep=0ex 
\belowrulesep=0ex 
\small
\setlength{\tabcolsep}{0.1cm}          
\begin{tabular}{c|cccc|c}
\toprule \rule{0pt}{1.0EM}
\# of clusters $\{k_m\}_{m=1}^M$ & Photo & Art. & Cartoon & Sketch & Avg.
\\ \midrule \rule{0pt}{1.0EM}

\{7\} & 65.63 & 56.43 & 54.97 & 65.46 & 60.62 
\\ \rule{0pt}{1.0EM}



\{7,14\} & 70.30 & 58.06 & 53.54 & 65.79 & 61.92 
\\ \midrule \rule{0pt}{1.0EM}



\{7,14,28\} (Ours) & \textbf{70.74} & \textbf{60.71} & \textbf{57.01} & \textbf{67.34} & \textbf{63.95} 

\\ \bottomrule
\end{tabular}

\caption{Effectiveness for varying numbers of clusters.}
\label{table4}
\vspace{-5mm}
\end{table}

\subsubsection{Effectiveness for varying numbers of clusters.}
\Cref{table4} shows the effect of using different numbers of clusters on the PACS dataset. It demonstrates that multiple cluster numbers consistently outperform single cluster settings.
As numbers of clusters increases, the model learns domain-irrelevant common features with more diverse attributes,
As numbers of clusters increases, the model learns diverse common features from different perspectives, 
leading to enhanced feature representation and domain generalization.

\begin{figure}[t]
\centering
\subfloat[]{\includegraphics[width=0.43\columnwidth]{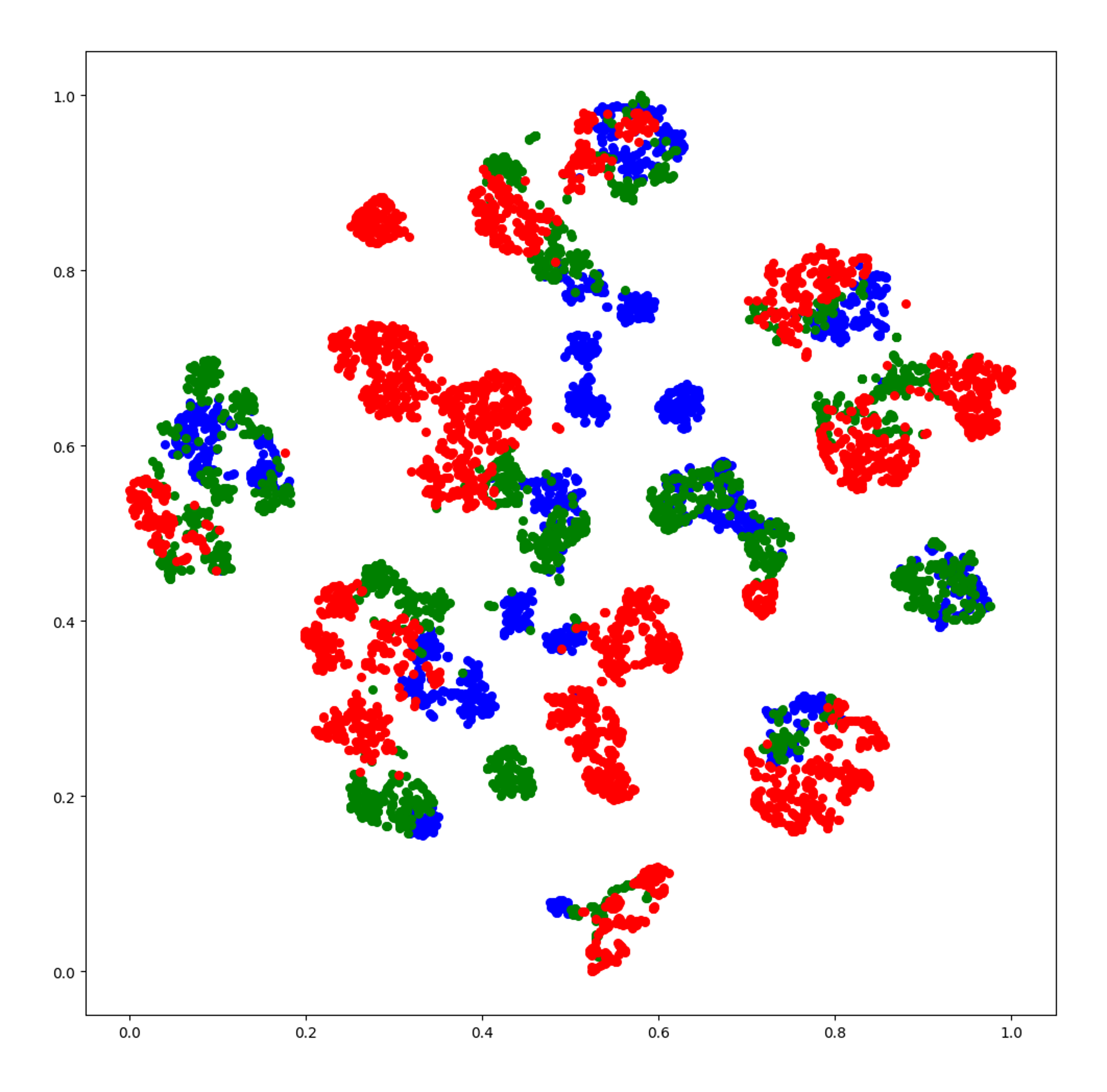}\label{41}}
\subfloat[]{\includegraphics[width=0.43\columnwidth]{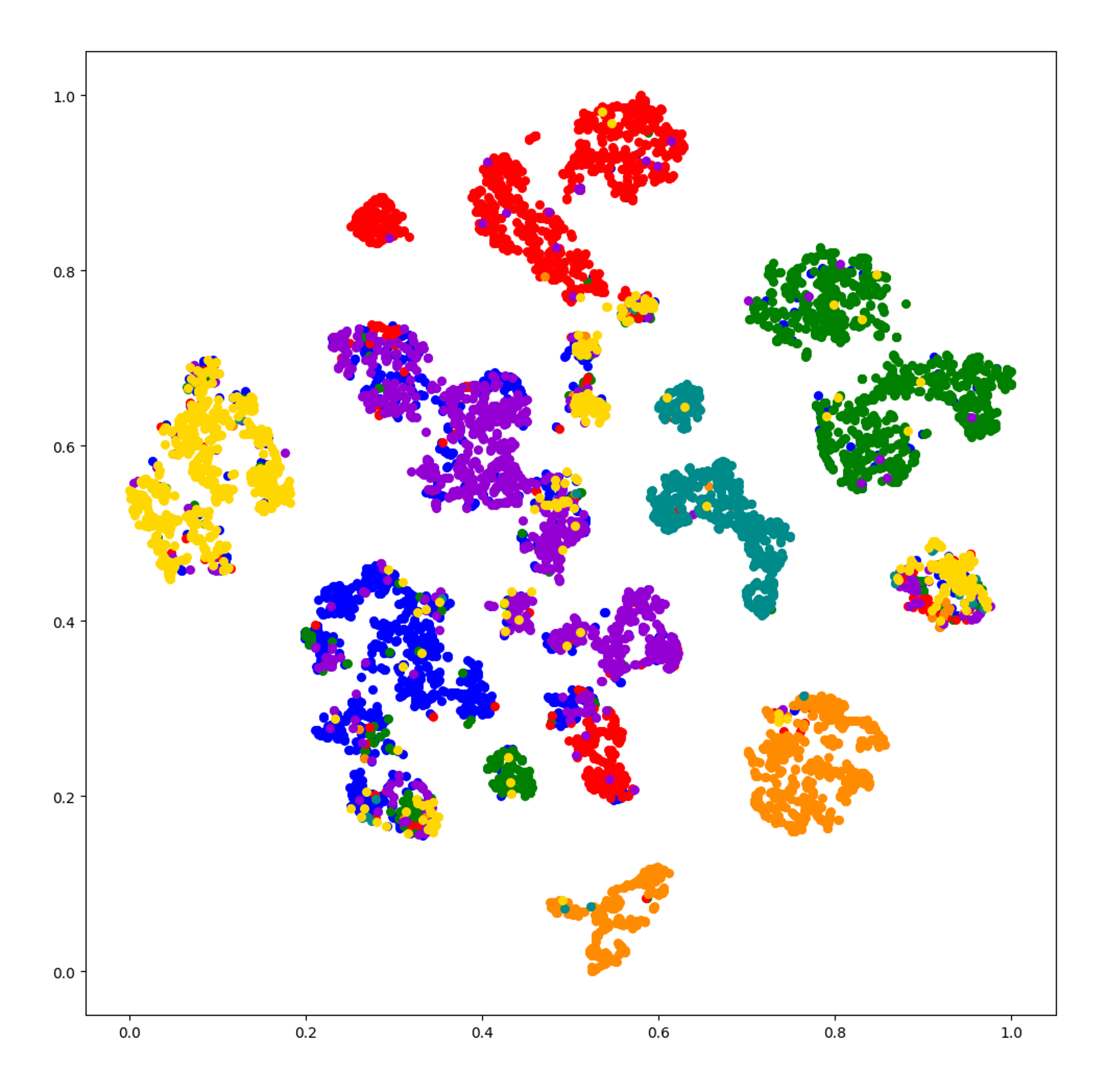}\label{42}}
\caption{Comparison of t-SNE visualization on PACS. The source train domains includes art painting, cartoon, sketch.
(a) Our proposed method on train samples with domain labels. (b) Our proposed method on train samples with class labels. T-SNE visualization for SimCLR is in Figure 1b.}
\label{figure5}
\vspace{-3mm}
\end{figure}

\subsubsection{T-SNE visualization.}
To better understand the effectiveness of our proposed method, we conducted a t-SNE visualization~\cite{tsne} on the PACS dataset, comparing it with SimCLR. In Figure 1b, SimCLR's representations show clear domain-based clustering, but less distinct class separation, indicating that SimCLR mainly captures domain-relevant features. In contrast, \Cref{figure5} shows the results of our method. Our method’s representations achieve better class discrimination while maintaining domain separation. It indicates that our method effectively captures domain-irrelevant common features. These results demonstrate that our method is effective to learn domain-irrelevant common features in UDG.

\begin{figure}[t]
  \centering
   \includegraphics[width=0.95\linewidth]{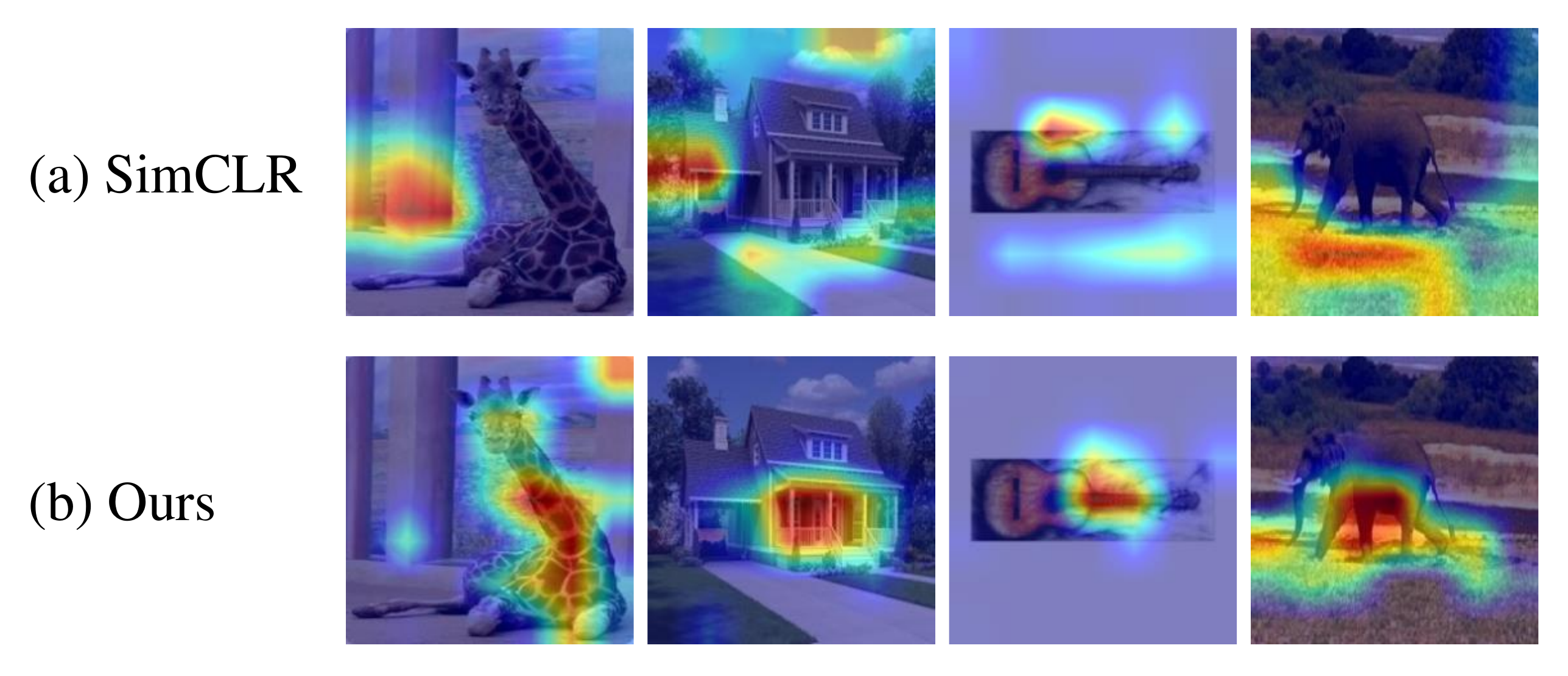}

   \caption{Comparison of Grad-CAM visualization on PACS.}
   \label{figure6}
\vspace{-5mm}
\end{figure}

\subsubsection{Grad-CAM visualization.}
To analyze what the model has learned from the features, we performed
Grad-CAM visualizations~\cite{gradcam, gradcam2}. In Figure 6a, the CAMs of SimCLR show that the model's attention is spread across the entire image or focused on domain-relevant features (e.g., background, texture). In contrast, Figure 6b shows that the CAMs of our proposed method focus the model's attention on the objects themselves (e.g., giraffes, houses, guitars, and elephants) while mostly ignoring domain-irrelevant background features. This demonstrates that our approach effectively captures domain-irrelevant common features.

%% file: CameraReady/LaTeX/sections/5_conclusion.tex
\section{Conclusion}
We addressed the issues with existing UDG methods, where instance discrimination tasks suppress domain-irrelevant common features and strong assumptions reduce the diversity of common features. 
To overcome these limitations, we proposed DomCLP to enhance domain-irrelevant common features and generalize common features across multiple domains without relying on strong assumptions. The proposed method demonstrated superior performance on the PACS and DomainNet datasets.

\section{Acknowledgments}
This work was partly supported by Institute of Information \& communications Technology Planning \& Evaluation (IITP) grant funded by the Korea government (MSIT) (RS-2019-II190421, AI Graduate School Support Program(Sungkyunkwan University), 20\%), Institute of Information \& Communications Technology Planning \& Evaluation (IITP) grant funded by the Korea government (MSIT) (No.RS-2024-00360227, Developing Multimodal Generative AI Talent for Industrial Convergence, 20\%), Institute of Information \& communications Technology Planning \& Evaluation (IITP) grant funded by the Korea government (MSIT) (No.2022-0-01045, Self-directed Multi-modal Intelligence for solving unknown, open domain problems, 20\%), the National Research Foundation of Korea (NRF) grant funded by the Korea government (MEST) (RS-2024-00352717, 20\%), and the IITP(Institute of Information \& Coummunications Technology Planning \& Evaluation)-ITRC(Information Technology Research Center) (IITP-2024-RS-2024-00437633, 20\%).

%% file: CameraReady/LaTeX/sections/6_appendix.tex
\clearpage
\onecolumn

\section{\normalsize Supplementary Material for \\ 
\LARGE DomCLP: Domain-wise Contrastive Learning with Prototype Mixup\\for Unsupervised Domain Generalization}
\vspace{1cm}

\section{Details for Setting and Datasets}
Tables 5 and 6 present the data split details for the PACS and DomainNet datasets. In the PACS dataset, we use three source domains, excluding one target domain, following existing approaches. For example, if Photo is the target domain, a model is trained on the pretraining data that consists of Art painting, Cartoon, and Sketch. 
To evaluate the encoder's ability to extract common features, a classifier is trained on a subset of the pretraining data (e.g., 1\% label fraction), and then accuracy is measured using the test data from Photo. 
In the DomainNet dataset, we train a model on three source domains and use the remaining domains as target domains. For instance, when the model and classifier are trained on pretraining data from Painting, Real, and Sketch, the target accuracies are measured using the test data from Clipart, Infograph, and Quickdraw, respectively.


\begin{table}[h]
\centering
\aboverulesep=0ex 
\belowrulesep=0ex 
\setlength{\tabcolsep}{0.3cm}          
\renewcommand{\arraystretch}{1.2}       
\begin{tabular}{c|cccc|c}
\toprule \rule{0pt}{1.0EM}
Phase & Photo & Art painting & Cartoon & Sketch & All.
\\ \midrule \rule{0pt}{1.0EM}
Pretraining & 1499 & 1840 & 2107 & 3531 & 8977
\\ \rule{0pt}{1.0EM}
Validation & 171 & 208 & 237 & 398 & 1014
\\ \midrule \rule{0pt}{1.0EM}
Test & 1670 & 2048 & 2344 & 3929 & 9991
\\ \bottomrule
\end{tabular}
\caption{The split details on PACS dataset for UDG. }
\label{table5}
\vspace{-5mm}
\end{table}

\begin{table}[h]
\centering
\aboverulesep=0ex 
\belowrulesep=0ex 
\setlength{\tabcolsep}{0.3cm}          
\renewcommand{\arraystretch}{1.2}       
\begin{tabular}{c|cccccc|c}
\toprule \rule{0pt}{1.0EM}
Phase & Painting & Real & Sketch & Clipart & Infograph & Quickdraw & All.
\\ \midrule \rule{0pt}{1.0EM}
Pretraining & 5305 & 9896 & 3901 & 3504 & 4804 & 9000 & 36410
\\ \rule{0pt}{1.0EM}
Validation & 600 & 1111 & 444 & 400 & 544 & 1000 & 4099
\\ \midrule \rule{0pt}{1.0EM}
Test & 5905 & 11007 & 4345 & 3904 & 5348 & 10000 & 40509
\\ \bottomrule
\end{tabular}
\caption{The split details on DomainNet dataset for UDG.}
\label{table6}
\end{table}

\section{Comparison of Representation Similarity Matrices}
\begin{figure}[!h]
\centering
\includegraphics[width=0.51\linewidth]
{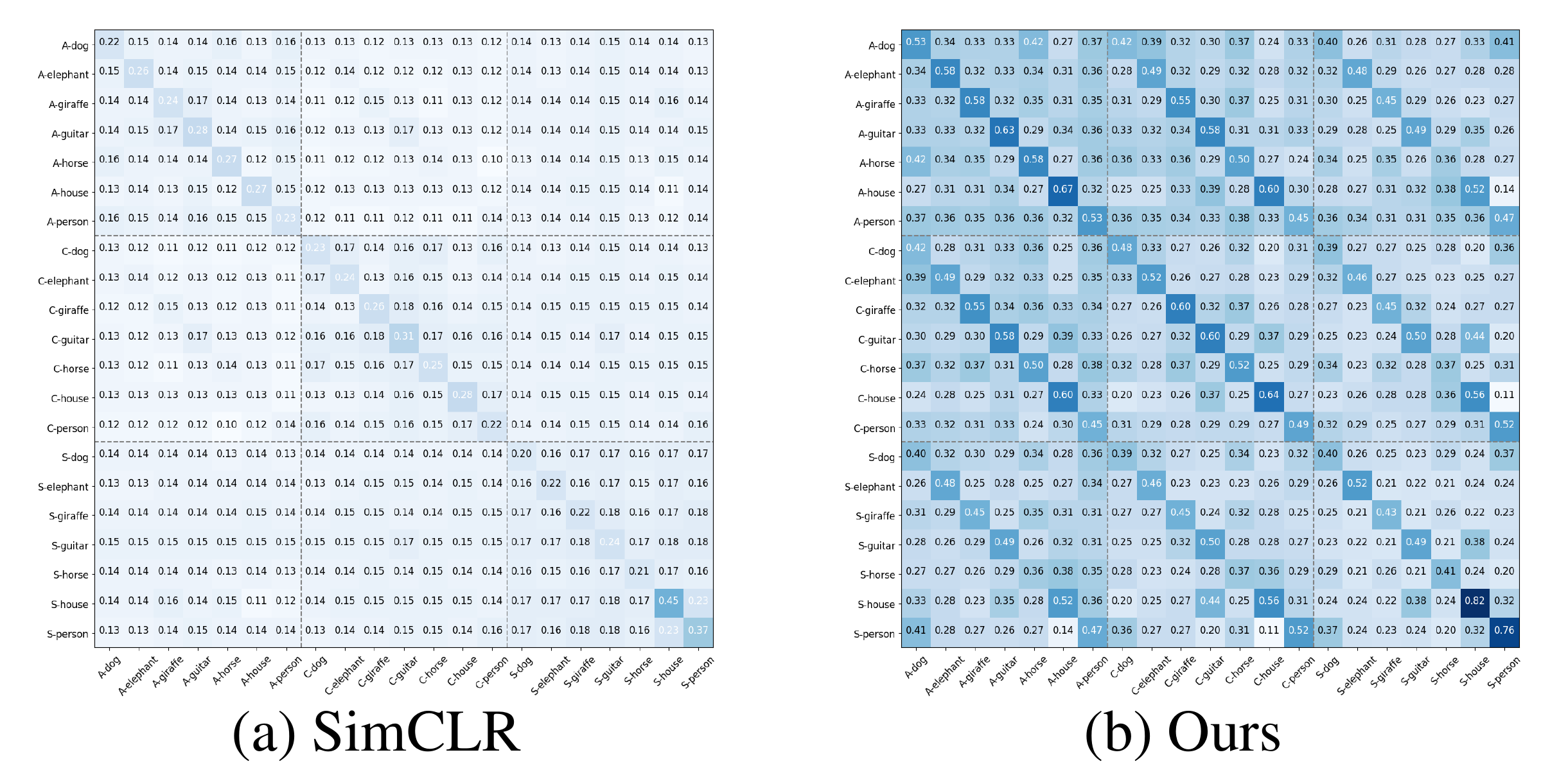}

\caption{Comparison of representation similarity matrices. Each cell in the matrix represents the average representation similarity between different domains and classes. The model is trained on source domain datasets consisting of Art painting, Cartoon, and Sketch domains.}
\label{figure7}
\end{figure}

\Cref{figure7} compares the representation similarity matrices obtained using SimCLR (a) and our proposed method (b) on the PACS dataset. Each cell in the matrix represents the average representation similarity between different domains and classes. 
In Figure 7a, the diagonal elements of the matrix are highlighted, while similarities between different domains are low. 
It indicates that SimCLR's representations mainly capture domain-relevant features. 
In contrast, with our proposed method, the average similarities between different domains but the same class are significantly higher. 
Furthermore, class-wise similarities of ours are obviously higher compared to those of SimCLR.
These results demonstrate that our method effectively learns domain-irrelevant common features in UDG.

\section{Algorithm for the Proposed Method}
The procedure for the proposed method is described in \Cref{algorithm1}. 
For $L_\text{pmix}$, prototypes for each domain are extracted in Lines 2$-$5. In \Cref{algorithm_line2}, representations $Z$ for all samples $X$ are extracted, with $Z$ dimensions as $N_S$ (number of samples) $\times$ $D_\text{proj}$ (projection dim).
\Cref{algorithm_line3} involves extracting prototypes with multiple clusters, and in \Cref{algorithm_line4}, we select samples with the same domain label and perform K-means clustering on each domain separately.
In Lines 8-14, losses of domain-wise contrastive learning, prototype mixup learning, and prototypical contrastive learning are extracted. 
Consequently, by learning with these losses, the model enhances and generalizes domain-irrelevant common features without relying on strong assumptions.
\begin{algorithm}[h]
    \SetKwInOut{Input}{Input}
    \SetKwInOut{Output}{Output}
    \SetKwInOut{Parameter}{Parameter}
    
    \Input{model encoder \( \theta \), projection head \( \psi \), multi-domain source datasets $S = \{(x_i, y^d_i)\}_{i=1}^{N_S}$}
    
    \Parameter{the number of clustering results $M$, numbers of clusters $K = \{k_1, \cdots , k_M \}$}
    
    \While{e \textless\ epochs}
    {
        $\mathcal{Z} = \psi(\theta(X))$ \textcolor{teal}{\it \# \textit{Extracting representations from all samples}} \nllabel{algorithm_line2} \\
        \For{$m\leftarrow 1$ \KwTo $M$ \nllabel{algorithm_line3}}
        {
            $\mathcal{P}^m \leftarrow \textit{Extracting Prototypes}(\mathcal{Z},\mathcal{Y^D}, k_m)$ \nllabel{algorithm_line4} \\
            
            \textcolor{teal}{\it \# \textit{K-means clustering on each domain separately with a number of clusters $k_m$}} \\
        }
        From $\mathcal{X}$, draw a mini-batch $\{(x_b, y^{d}_b, p^1_b, \cdots, p^M_b); b \in (1, ..., B), p^m \in \mathcal{P}^m$\} \\
        \For{$b\leftarrow 1$ \KwTo $B$}
        {
            $L_{dcon} \leftarrow \mathrm{DCon}(x_b, y^{d}_b)$ with \Cref{equation4} \textcolor{teal}{\it \# \textit{DCon: \underline{D}omain-wise \underline{Con}trastive Learning}} \\
            \For{$m\leftarrow 1$ \KwTo $M$}
            {
                $L^m_{pmix} \leftarrow \mathrm{PMix}(x_b, p^m_b)$ with \Cref{equation5} \textcolor{teal}{\it \# \textit{PMix: \underline{P}rototype \underline{Mix}up Learning}} \\
                
                $L^m_{pcl} \leftarrow \mathrm{PCL}(x_b, p^m_b)$ with \Cref{equation6} \textcolor{teal}{\it \# \textit{Prototypical Contrastive Learning}} \
            }
            $L_{pmix}, L_{pcl} \leftarrow \frac{1}{M} \sum_{m=1}^M L^m_{pmix}, \frac{1}{M} \sum_{m=1}^M L^m_{pcl}$ \\
            $L = L_{dcon} + L_{pmix} + L_{pcl}$ \\
            
            $\theta, \psi \leftarrow \mathrm{SGD}(L,\theta,\psi)$ \textcolor{teal}{\it \# \textit{Update model parameters by minimizing $L$}}
        }
     }
     
\caption{\textbf{Dom}ain-wise \textbf{C}ontrastive \textbf{L}earning with \textbf{P}rototype Mixup (DomCLP)}
\label{algorithm1}
\end{algorithm}

\vspace{-0.5cm}
\section{Comparison of Feature Diversity}
Existing UDG approaches have attempted to generalize common features across multiple domains through feature alignment. These methods merge the feature manifolds of each domain into a single manifold by relying on strong assumption-based feature alignments. However, these strong assumptions may result in learning biased features or reducing the diversity of the common features.
To verify that strong assumption-based alignments reduce feature diversity, we analyze the condition number of the representation matrix. 
The representation matrix is composed of representations from all samples, and its condition number serves as an indicator of how well the feature space is covered across various dimensions.
The condition number of the representation matrix is defined as follows:
\begin{equation}
\kappa(R) = \|R\|\|R^{-1}\| = \frac{\lambda_1(R)}{\lambda_{n}(R)} 
\label{equation8}
\end{equation}
\noindent where $R$ indicates the representation matrix, with $\lambda_1 (R)$ as its largest eigenvalue and $\lambda_n (R)$ as its $n$-th largest eigenvalue. 
The condition number is defined by the ratio of these eigenvalues. 
If the representation captures feature information across various dimensions, the eigenvalues are spread out, resulting in a small condition number, and vice versa.
As shown in \Cref{table7}, existing methods that reduce the diversity of domain-irrelevant common features have larger condition numbers because they tend to learn only a limited set of common features. In contrast, our method learns diverse common features, resulting in the lowest condition number. This demonstrates that our approach effectively captures common features across multi-domain data.

\begin{table}[!h]
\centering
\aboverulesep=0ex 
\belowrulesep=0ex 
\setlength{\tabcolsep}{0.1cm}          
\begin{tabular}{c|cccc|c}
\toprule \rule{0pt}{1.0EM}
Method & Photo & Art. & Cartoon & Sketch & Avg.
\\ \midrule \rule{0pt}{1.0EM}

BrAD & 13.94 & 10.63 & 99.85 & 15.60 & 35.01
\\ \rule{0pt}{1.0EM}

BSS & 19.32 & 13.06 & 28.95 & 10.31 & 17.91 
\\ \midrule \rule{0pt}{1.0EM}

Ours & \textbf{2.84} & \textbf{3.11} & \textbf{2.50} & \textbf{3.13} & \textbf{2.90} \\ \bottomrule
\end{tabular}

\caption{Condition number for the representation matrix. The hyperparameter $n$ for the condition number is set to 10.}
\label{table7}
\end{table}